\documentclass[10pt,journal,compsoc]{IEEEtran}
\ifCLASSOPTIONcompsoc
  \usepackage[nocompress]{cite}
\else
  \usepackage{cite}
\fi

\ifCLASSINFOpdf
\else
\fi

\usepackage{amsmath,amsfonts}
\usepackage{algorithmic}
\usepackage{algorithm}
\usepackage{array}
\usepackage[caption=false,font=normalsize,labelfont=sf,textfont=sf]{subfig}
\usepackage{textcomp}
\usepackage{stfloats}
\usepackage{url}
\usepackage{verbatim}
\usepackage{graphicx}
\usepackage{float}
\usepackage{color}

\usepackage{makecell}

\begin{document}

%
\title{MO-MIX: Multi-Objective Multi-Agent Cooperative Decision-Making With \\Deep Reinforcement Learning}
%
%
%
%

\author{Tianmeng Hu,
        Biao Luo, ~\IEEEmembership{Senior Member, IEEE},
        Chunhua Yang, ~\IEEEmembership{Fellow, IEEE},\\
        Tingwen Huang, ~\IEEEmembership{Fellow, IEEE}
\IEEEcompsocitemizethanks{\IEEEcompsocthanksitem T. Hu, B. Luo and C. Yang are with the School of Automation, Central South University, Changsha, 410012, China.\protect\\
E-mail: \protect\url{tianmeng0824@163.com}, \protect\url{biao.luo@csu.edu.cn}, \protect\url{ychh@csu.edu.cn}
\IEEEcompsocthanksitem T. Huang is with the Texas A\&M University at Qatar, Doha, 23874, Qatar. E-mail: \protect\url{tingwen.huang@qatar.tamu.edu}}
}

%
%

\markboth{IEEE Transactions on Pattern Analysis and Machine Intelligence}%
{Shell \MakeLowercase{\textit{et al.}}: Bare Demo of IEEEtran.cls for Computer Society Journals}
%

\IEEEpubid{\makebox[\columnwidth]{\hfill 10.1109/TPAMI.2023.3283537~\copyright~2023 IEEE}%
\hspace{\columnsep}\makebox[\columnwidth]{Published by the IEEE Computer Society\hfill}}


\IEEEtitleabstractindextext{%
\begin{abstract}
  Deep reinforcement learning (RL) has been applied extensively to solve complex decision-making problems. In many real-world scenarios, tasks often have several conflicting objectives and may require multiple agents to cooperate, which are the multi-objective multi-agent decision-making problems. However, only few works have been conducted on this intersection. Existing approaches are limited to separate fields and can only handle multi-agent decision-making with a single objective, or multi-objective decision-making with a single agent. In this paper, we propose MO-MIX to solve the multi-objective multi-agent reinforcement learning (MOMARL) problem. Our approach is based on the centralized training with decentralized execution (CTDE) framework. A weight vector representing preference over the objectives is fed into the decentralized agent network as a condition for local action-value function estimation, while a mixing network with parallel architecture is used to estimate the joint action-value function. In addition, an exploration guide approach is applied to improve the uniformity of the final non-dominated solutions. Experiments demonstrate that the proposed method can effectively solve the multi-objective multi-agent cooperative decision-making problem and generate an approximation of the Pareto set. Our approach not only significantly outperforms the baseline method in all four kinds of evaluation metrics, but also requires less computational cost.
\end{abstract}

}

\maketitle

\IEEEdisplaynontitleabstractindextext

%
\IEEEpeerreviewmaketitle

\IEEEraisesectionheading{\section{Introduction}\label{sec:introduction}}

%
%
%
%
\IEEEPARstart{M}{ulti-agent} reinforcement learning (MARL) methods \cite{gupta2017cooperative, lowe2017multi, Peter2018, foerster2018counterfactual, rashid2018qmix, sun2021reinforcement, son2019qtran, zhang2021learn} solve multi-agent decision problems by reinforcement learning \cite{watkins1992q, mnih2015human, lillicrap2016continuous, schulman2015trust, yang2018hie, schulman2017proximal, dilokthanakul2019feature, dong2021dynamical, shi2020face}, typically using a reward function to train agents to cooperate or compete on a specific task. The optimization objective is determined by a reward function, which assigns a reward to each decision made by the agent.
However, many real-world problems have multiple conflicting objectives. For example, an autonomous driving system must consider two objectives: passenger comfort and vehicle speed. If efficiency is critical, vehicle speed must be increased, which may result in more hard braking or lane changing behaviors. On the other hand, if passenger comfort is important, the vehicle should travel smoothly, which means a reduction in average speed.
Conflicting objectives cannot be optimal at the same time. Therefore, a policy must trade-off between different objectives. In other words, we assign a weight representing the importance to each objective, which reflects the preference over the objective. 
For multi-objective decision-making problems, the common solution converts the multi-objective problem into a single-objective one by constructing a synthetic reward function, e.g., using a weighted sum of the rewards for several objectives \cite{schulman2017proximal, duan2016benchmarking, lowe2017multi}. In this way, a policy can be found by traditional RL methods. However, such approaches have following weaknesses: 1) only a single policy can be found, which is optimized for a fixed preference over the objectives, and 2) it is challenging to find the optimal weights manually. Another solution is the multi-objective reinforcement learning (MORL), which can learn policies for different preferences and find a Pareto set approximation. The MORL framework has distinct advantages over the single-objective RL: 1) different policies can be found to adapt to different preferences; 2) the synthetic reward function is no longer required.

Most recently, extensive research has been conducted in the field of single-agent multi-objective reinforcement learning \cite{van2014multi, reymond2019pareto, abdelfattah2019intrinsically, mossalam2016multi, abels2019dynamic, chen2019meta, xu2020prediction, yang2019generalized}, but none of these approaches can solve the multi-agent cooperative decision-making problem. This is because the single-agent RL algorithms cannot address: 1) the non-stationary environment problem: for a given agent, all other agents become part of the environment, and the policies of those agents are constantly changing during the training process, which significantly increases the dynamic and instability of the environment; 2) the partially observable problem \cite{lowe2017multi}: agents can only obtain partial information of the environment and cannot get the observations, actions and rewards of other agents; 3) the credit assignment problem \cite{Peter2018, rashid2018qmix}: the value function estimated using the joint rewards cannot evaluate the contribution of each agent, and an agent with a bad policy may be spuriously rewarded because of other agents' behaviors.

To address these challenges, in this paper, we introduce a novel MOMARL algorithm, which is used to learn a decision-making model that can generalize to different preferences over the objectives. Further, by inputting different preferences, we can get a dense and high-quality set of policies that constitute a Pareto set approximation. The proposed approach is not only able to address the multi-agent decision-making problem, but also generates corresponding optimal policies for different preferences. In summary, our main contributions are as follows.

\begin{enumerate}
  \item{We propose a novel multi-objective multi-agent cooperative decision-making method, called MO-MIX, which is able to generate various policies based on the input preferences and finally achieve a dense and high-quality Pareto set approximation. To the best of our knowledge, this is the first multi-objective reinforcement learning approach that is applicable to multi-agent systems and yields high-quality non-dominated sets. }
  \item{We propose an exploration guide approach. The exploration direction of the algorithm is guided during the training process, which can improve the uniformity of the final Pareto set approximation. }
  \item{We evaluate our algorithm in OpenAI's multi-agent particle environment. For comparison, we constructed an outer-loop MOMARL algorithm using QMIX as a baseline. The experimental results show that the proposed MO-MIX method generates higher quality non-dominated sets with distinct advantages in all four kinds of evaluation metrics. In addition, our method has significant efficiency advantage and requires much less computational costs. }
\end{enumerate}

The rest of this paper is organized as follows. Firstly, we review the related work in Section \ref{sec2}, including MARL, MORL and MOMARL. Then, we briefly introduces the background of multi-objective decision-making and propose a formalization of the MOMARL problems in Section \ref{sec3}. The proposed MO-MIX method is presented in detail in Section \ref{sec4}. Finally, experimental results and analysis are given in Section \ref{sec5} and our work is concluded in Section \ref{sec6}.

\section{Related Work}
\label{sec2}
In recent years, outstanding progress has been achieved in deep reinforcement learning, with a significant increase in applications\cite{ shi2020face, luo2020end, rao2021visual, xu2019predicting, dong2021dynamical} and methods\cite{cao2021weak, dilokthanakul2019feature, chen2022domain, yang2018hie, keneshloo2020deep, zhang2021learn, sun2021reinforcement}. Many recent works have turned their focus to multi-agent systems or multi-objective decision-making, which are two challenging problems in the reinforcement learning domain. We will introduce some relevant works in the following.

\subsection{Multi-agent Reinforcement Learning}
MARL is a series of reinforcement learning methods that consider multi-agent learning scenarios. In contrast to single-agent systems, multi-agent systems have several serious challenges, e.g., the non-stationary of environment, the partially observable environments, the curse of dimensionality of the state and joint action space, the credit assignment challenge, etc. In this paper, we focus on the domain of multi-agent cooperation, where several agents must cooperate to accomplish tasks. An agent must cooperate with other agents by considering both the state of the environment and the behavior of other agents.
The independent Q-learning (IQL) algorithm \cite{tan1993multi} is a simple MARL framework. IQL treats each agent independently and each agent can get local observations. The reinforcement learning algorithm is applied to each agent independently. Although IQL can address some problems\cite{gupta2017cooperative, tampuu2017multiagent}, it has the non-stationarity of environment issue\cite{oroojlooyjadid2019review}.
One solution to this problem is using a framework called centralized training with decentralized execution. For example, Lowe et al. \cite{lowe2017multi} proposed the multi-agent deep deterministic policy gradient (MADDPG) algorithm, in which agents select actions using an actor based on partial observations, while estimating action values using a critic based on global observations. In addition, there are several works that improve the MADDPG, e.g., the multi-actor-attention critique \cite{iqbal2019actor} and the counterfactual multi-actor policy gradient \cite{foerster2018counterfactual} algorithms.

The credit assignment is another major challenge in multi-agent cooperation. Since agents share rewards, a poorly performing agent may get high rewards because of behaviors of other agents, which are spurious rewards and may lead to lazy agents. To address this issue, Sunehag et al. \cite{Peter2018} proposed the value-decomposition network (VDN).
The idea is to decompose the value function of the team into several sub-value functions, which are used as the basis for each agent to select actions. Such a design makes the agents not share the same value function, which alleviates the credit assignment problem.
To be specific, the VDN uses the sum of the partial action-value functions of each agent as the joint action-value function:
\begin{equation}
  \label{vdn}
  Q^{tot}\left( \boldsymbol{\tau}, \boldsymbol{a} \right) = \sum_{i=1}^{n} Q^{i}\left( \tau^{i}, a^{i} \right),
\end{equation}
where $Q^i$ is the value function of agent $i$. $\tau$ and $\boldsymbol{\tau}$ are the partial and global observation histories, respectively. $\boldsymbol{a}$ is the joint action, which consists of the action $a^{i}$ of each agent $i$.

Further, Rashid et al. \cite{rashid2018qmix} proposed the monotonic value function factorization (QMIX) algorithm, which is an improvement of VDN.  The VDN method assumes that the total value function is equal to the sum of all sub value functions, which is a too strong constraint that limits the performance of the algorithm. In contrast, the QMIX algorithm uses a mixing network to decompose the value function and achieves better performance. In addition, QMIX uses hypernetworks to make the output of the network satisfy the following monotonicity constraint:
\begin{equation}
  \label{qmix}
  \frac{\partial Q^{tot}}{\partial Q^{i}} \geq 0.
\end{equation}
Although QMIX can cover many multi-agent problems, it has trouble in some cooperative problems with non-monotonic reward. To address this issue, the Weighted QMIX \cite{rashid2020weighted} and QTRAN \cite{son2019qtran} algorithms have improved QMIX in different ways.
Recently, MARL has also shown potential in many real-world problems. Xu et al. \cite{Xu2021multiagent} proposed a distributed transmission scheme based on MARL in a collaborative cloud-edge architecture, and proposed a hybrid learning framework by extending the actor-critic model to achieve centralized training and decentralized execution. This approach shows superior performance in terms of network delay and quality of service. Liu et al. \cite{Liu2021CMIX} proposed the CMIX algorithm, which combines the constrained reinforcement learning (CRL) and CTDE frameworks. This approach considers both peak and average constraints, modifies the reward function to handle peak constraint violations, and converts the problem under the average constraint to a max-min optimization problem.The CMIX algorithm can optimize the RL reward while satisfying both constraints. 

\subsection{Multi-objective Reinforcement Learning}
Real-world decision-making problems often involve several conflicting objectives. MORL extends the current RL methods to be able to address two or more objectives. Current MORL methods can be classified into two categories: the single-policy and multi-policy methods \cite{liu2014multiobjective, roijers2013survey}. The single-policy methods convert a multi-objective decision-making problem into a single-objective one and attempt to find a single policy by using the traditional RL methods. For example, Schulman et al. \cite{schulman2017proximal} designed a synthetic reward function for robot control by selecting a weight for each objective manually.
These methods are easy to implement and requires less computational costs but has the following problems: 1) it requires some prior information to decide the weights of different objectives; 2) it is difficult and nonintuitive to find the best weights; 3) only a single policy can be found. In many scenarios, we would expect to find multiple policies to address different requirements, such as energy efficiency or high performance.

The multi-policy methods allow finding multiple solutions so that users can choose suitable solutions to satisfy various preferences. In other words, the multi- policy methods attempt to approximate the true Pareto frontier by learning multiple policies \cite{barrett2008learning}, which can intuitively present the users with the trade-off information among the conflicting objectives.
Mossalam et al. \cite{mossalam2016multi} proposed the scalarized Q-learning method that extends the deep Q-network \cite{mnih2015human} to address the MORL problems. The vector value function is scalarized by computing the inner product of the value function and the preference. The algorithm uses an outer loop to search for preferences and eventually generates a series of policies.
Chen et al. \cite{chen2019meta} proposed a MORL method based on meta-learning \cite{finn2017model} (Meta- MORL). A meta-policy is first trained and used as the best initial policy. Next, the meta-policy is fine-tuned by running a small number of gradient updates to generate solutions with different preferences.
Xu et al. \cite{xu2020prediction} proposed the prediction-guided evolutionary learning algorithm that guides the optimization direction of several parallel RL tasks through a prediction model. This approach maintains an nondominated set during the learning phase and approximates the Pareto frontier by reinforcement learning.
Abels et al. \cite{abels2019dynamic} proposed the multi-objective Q-network method that use a single neural network to represent the value function of the entire preference space. Yang et al. \cite{yang2019generalized} improved this method to achieve better and faster learning by using the envelope Q-learning algorithm.

It is necessary to give a brief description of the difference between MORL and CRL. CRL optimizes the RL objective under one or more hard constraints to find a single optimization policy that satisfies all constraints. In contrast, MORL has no constraints. As discussed previously, some MORL methods try to find a single policy that achieves a satisfying balance between multiple conflicting objectives. In this case, the problem can be transformed into a CRL problem. Other MORL methods, the multi-policy methods, try to find a set of policies, which is the Pareto approximation set for the multi-objective decision-making problem. Multi-policy MORL has several advantages: 1) It provides multiple solutions for different objective preferences, rather than a single solution. In practice, this allows the user to freely choose satisfactory solutions from a solution set. 2) There are situations where natural constraints do not exist. Consider a robot control problem with two objectives, speed and energy conservation. The user may view energy conservation as the more important objective, but does not know how to set the energy consumption constraint. An inappropriate constraint setting may result in too low a speed, leading to an unsatisfactory solution. 3) The preferences or constraints on the objectives may change. For CRL, this requires changing the constraints and retraining the agents.

\subsection{Multi-objective Multi-agent Reinforcement Learning}
Several works have been conducted to address the multi-objective multi-agent reinforcement learning problems. Khamis et al. \cite{khamis2014adaptive} applied RL methods to the traffic signal control problem, which is a MOMARL problem. Also, Mannion et al. \cite{mannion2016multi} used RL methods to address the dynamic economic emissions dispatch problem. Both above works converted the multi-objective problem into a single-objective one by scalarizing the reward signal vector, which leads to a single solution. Van Moffaert et al. \cite{van2014novel} proposed an adaptive weight selection algorithm for MORL that can better cover the Pareto front. Mannion et al. \cite{mannion2018reward} applied several reward shaping methods to the MOMARL problem and discussed the theoretical implications.
Robinson et al. \cite{Robinson2021multi} proposed a multi-agent task planning method for robot task planning, which treats multiple objectives as constraints to find a single optimal solution that satisfies all constraints. Tittaferrante et al. \cite{Tittaferrante2022multi} used multiadvisor reinforcement learning to solve the multi-agent multi-objective smart home energy control problem. This approach has better performance than the single-objective RL, but still has some limitations: 1) each agent applies an independent DQN algorithm instead of a MARL algorithm, which may lead to reduced stability and effectiveness of the algorithm in a more complex multi-agent environment; 2) this approach cannot provide multiple non-dominated solutions corresponding to different objective preferences. 

The above works have explored the MOMARL problem from different perspectives. However, the research in this intersection is still not deep enough. Current works are not able to address complex cooperative decision-making problems with continuous or high-dimensional state space. Also, most of them simply utilize the single-policy methods. In contrast, our work has several improvements: 1) our approach generates multiple policies and approximates the Pareto frontier by a single model, thus the trained model can execute the optimal policy according to the given preferences; 2) our approach extends the advanced deep RL method to the MOMARL domain for the first time, and can solve decision-making problems with continuous state space; and 3) compared to the single-agent MORL methods, our approach can handle the non-stationarity of environment and the credit assignment challenge in multi-agent systems.

\section{Preliminaries}
\label{sec3}
In this section, we present some concepts of multi-objective decision-making briefly. Then, we introduce the multi-objective Dec-POMDP, which is a formalization of MOMARL.

\subsection{Multi-objective Decision-making}
For reinforcement learning, a multi-objective decision-making problem can be stated as:
\begin{multline}
  \label{moo}
  \max _{\pi\left(s\right)} {F}\big( \pi\left(s\right) \big)= \\
  \max _{\pi\left(s\right)}
  \Big[f_{1}\big(\pi\left(s\right)\big), f_{2}\big(\pi\left(s\right)\big), \ldots, f_{m}\big(\pi\left(s\right)\big)\Big],
\end{multline}
where $\pi\left(s\right)$ is the policy and $ f_{1}, f_{2}, \ldots, f_{m} $ are the objective functions of the $m$ objectives.

{\bf{Pareto optimality}} \cite{zhang2007moea}. Policy $\pi_1\left(s\right)$ dominates policy $\pi_2\left(s\right)$ if and only if $ f_{i}\big(\pi_1\left(s\right)\big) \geq f_{i}\big(\pi_2\left(s\right)\big) $ for every $ i \in \left( 1, \ldots, m \right) $ and $ f_{j}\big(\pi_1\left(s\right)\big) > f_{j}\big(\pi_2\left(s\right)\big) $ for at least one $ j \in \left( 1, \ldots, m \right) $. If a policy is not dominated by any other policies, then it is a Pareto optimal policy. Any improvement of such a policy in one objective will lead to regression in at least one other objectives.

{\bf{Pareto set}}. No single policy can optimize all the objectives in a multi-objective decision-making problem. Ideally, the Pareto set, which is the set of all Pareto optimal policies, is required. However, in real-world applications, the true Pareto frontier is usually not available. Instead, a set of non-dominated policies is available, which is an approximation of the true Pareto set. 
A non-dominated set is composed of multiple policies. In a non-dominated set, no policy can outperform other policies on all objectives. There are no advantages or disadvantages among these policies, but only different preferences over the objectives.  
Therefore, the purpose of multi-objective decision-making is to achieve a set of high-quality policies which can approximate the Pareto frontier in the objective space.

\subsection{Multi-objective Dec-POMDP}
A cooperative multi-agent decision-making problem can be described as a decentralized partially observable Markov decision process (Dec-POMDP) \cite{oliehoek2008optimal}, which is an extension of the Markov decision process (MDP) \cite{bellman1957markovian}. We extend the Dec-POMDP to the multi-objective domain, i.e., the multi-objective Dec-POMDP consisting of a tuple $ \mathcal{G}=\langle\mathcal{S}, \mathcal{A}, \mathcal{P}, \mathbf{r}, \mathcal{Z}, \mathcal{O}, N, \gamma\rangle $, where $ s \in \mathcal{S} $ is the global state of the environment, $\mathcal{A}$ is the action set for each agent, $ \mathbf{r}=\left[r_{1}, \ldots, r_{m}\right]^{\top} $ is a vector reward function corresponding to $m$ objectives, and $\mathcal{P}$ is a probabilistic transition function.
Note that the reward vector is the team reward.
In the partially observable environments, an agent is allowed to get a partial observation $ o \in \mathcal{O} $ from the observation function $\mathcal{Z}$, which is the basis for the decision making.
Each agent chooses an action $ a \in \mathcal{A} $ at each time step according to the policy $\pi\left(s\right)$. The actions chosen by $N$ agents form a joint action space $\boldsymbol{a} \in \mathcal{A}^{N}$. The environment then transforms to next state according to the transition function $ \mathcal{P}\left(s^{\prime} \mid s, \boldsymbol{a}\right) $, while returning a vector reward $ \mathbf{r}\left(s, \boldsymbol{a}\right) $. 
$\gamma$ is the discount factor. Future rewards are discounted by $\gamma \in [0,1]$, which is designed to strike a balance between immediate and long-term rewards.

The purpose of reinforcement learning is to find a policy $\pi$ that maximizes the expected return. For the state-action pair $ \left(s, \boldsymbol{a}\right) $ at time step $t$, the return is a discounted sum of future rewards:
\begin{equation}
  \label{Gt}
  \mathbf{G}_{t}=\sum_{t^{\prime}=t}^{\infty} \gamma^{t^{\prime}-t} \mathbf{r}\left(s_{t^{\prime}}, \boldsymbol{a}_{t^{\prime}}\right).
\end{equation}
The $\mathbf{G}$ is also a vector, where each element corresponds to the return on one objective.

\section{Our Approach}
\label{sec4}

In this section, we propose the MO-MIX method, an end-to-end RL approach that is designed to address the multi-objective multi-agent cooperative
decision-making, which can be formalized as a multi-objective DecPOMDP. Firstly, we introduce two components of the MO-MIX: the Conditioned Agent Network and Multi-objective Mixing Network. Secondly, we propose an exploration guide approach for improving the uniformity of the distribution of the final non-dominated solutions. Finally, we present the procedure of the MO-MIX algorithm.

\subsection{Conditioned Agent Network}

\begin{figure*}[tp]
  \centering
  \includegraphics[width=0.95\textwidth]{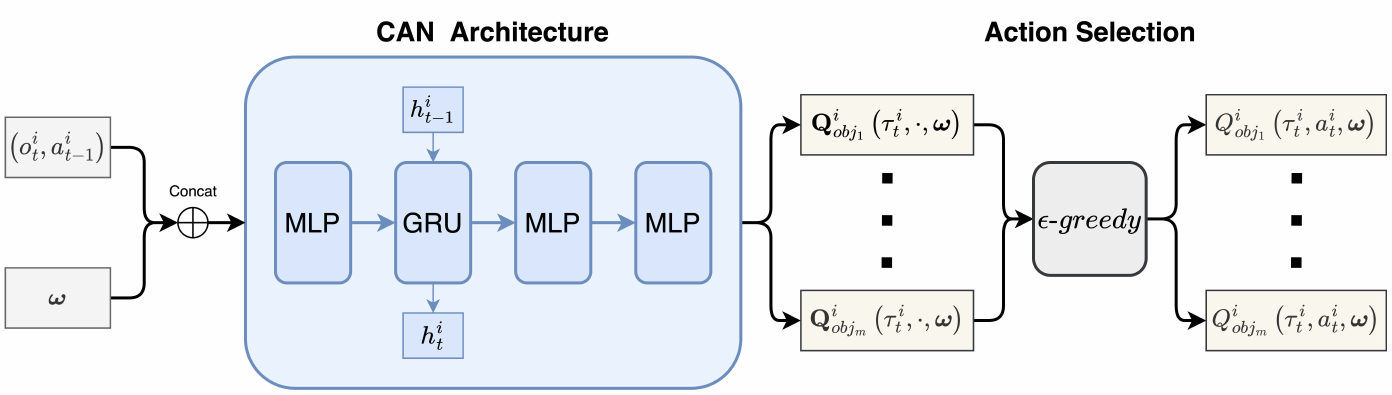}
  \caption{The architecture of the CAN. Each agent uses its respective CAN to estimate the partial multi-objective Q-function and to select the action. The inputs to a CAN are the partial observation of agent $i$, the action of the previous time step and the current preference $\boldsymbol{w}$. That is, we use the preference $\boldsymbol{w}$ as a condition for estimating the value function. A GRU layer is used to take advantage of the partial observation history. The output of the CAN is multi-objective Q-vectors for each optional action, each vector containing Q-values for $m$ objectives. The agents select actions independently based on the $\epsilon$-greedy policy. }
  \label{CAN}
\end{figure*}

\begin{figure*}[tp]
  \centering
  \includegraphics[width=0.95\textwidth]{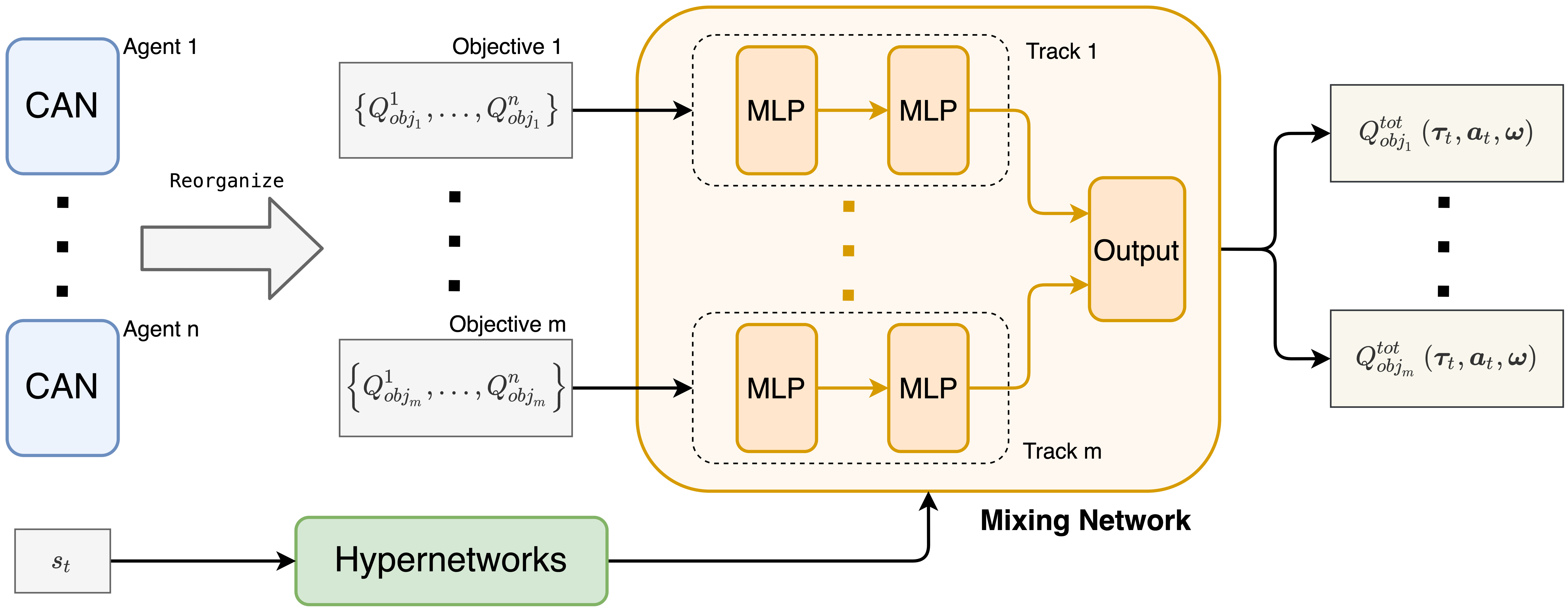}
  \caption{The architecture of the MOMN. The MOMN takes outputs of all CANs and reorganizes them according to different objectives. The MOMN is internally divided into $m$ parallel tracks corresponding to $m$ objectives. Each track has two MLP layers. Multiple track outputs are then concatenated as the outputs of the entire network, which are the multi-objective joint Q-values. The global state $s_{t}$ at time step $t$ is fed to several hypernetworks and used to generate weights and biases for the mixing network. }
  \label{MOMN}
\end{figure*}

The action-value function $Q\left(s,a\right)$ in reinforcement learning is the expected cumulative discounted return corresponding to a certain action $a$ taken at a certain state $s$:
\begin{equation}
  \label{Qfunction}
  Q(s, a)=\mathbb{E}_{\pi}\left[G_{t} \mid s_{t}=s, a_{t}=a\right].
\end{equation}
The optimal policy can be formulated by selecting the action that maximizes the value function. However, considering the multi-objective decision-making problem, a vector action-value function $\mathbf{Q}$ is required. We propose the Conditioned Agent Network (CAN) for estimating the vector function $\mathbf{Q}$, which consists of the value functions of all objectives.

The CAN consists of several Multilayer Perceptron (MLP) layers and a Gate Recurrent Unit (GRU) \cite{cho2014learning} layer. GRU is one kind of recurrent neural network that can better address sequential information.
Hausknecht et al. \cite{hausknecht2015deep} first combined RNN with RL methods, and Peter et al. \cite{Peter2018} introduced it to MARL. We choose GRU because it performs well and is easy to compute.
The output of GRU is not only related to the input of the current time step but also to the historical input. Therefore, the CAN has the ability to utilize the entire observation and action history of the agent, which compensates for the deficiency of local observations \cite{Peter2018}. In addition, considering the adaptation of dynamic weights is required, we connect two layers of MLP after GRU in order to improve the representation capability of the network.

In this work, we are not only to find one solution corresponding to a specific preference, but to find multiple non-dominated solutions, i.e., to find a Pareto set approximation. Therefore, we need to train the model using different preferences, thus increasing the diversity of the non-dominated set.
To do this, the preference vectors are used as part of the network input, thus a trained model produces the appropriate policy based on the input preference. Without this setting, the algorithm will output only a single policy corresponding to a specific preference. 
Each agent uses an independent CAN for the estimation of their respective action value functions. The inputs to the network are the observation and action information of the agent, along with a preference vector $\boldsymbol{\omega}$ representing the preference. The purpose of the input preference vector is to make the neural network estimate the vector Q-function with a specific weight as condition.
Since the dimensionality of the observation and action information is significantly larger than that of the preference vector, to avoid the preference information being ignored by the model, we replicate the preference vector $\boldsymbol{\omega}$ multiple times. 
For example, if there are two objectives, the original preference vector is a 2-dimensional vector, which we replicate to 14-dimensional and then concatenate with the state information as the input to the network.  

We consider the discrete action space $\mathcal{A}$, consisting of $n$ actions $a_{1},\ldots,a_{n}$. At each time step, a CAN outputs a vector $ \mathbf{Q}\left(\tau,\cdot,\boldsymbol{\omega}\right) $ for each optional action, based on the agent's observation history $\tau$. During the training process, the agent's behavior follows the $\epsilon$-greedy rule, i.e., random actions are selected with a probability $\epsilon$, otherwise the optimal action is chosen. The optimal action is defined based on the scalarization of the vector $\mathbf{Q}$, which means that the current weight $\boldsymbol{\omega}$ is used to weight the $\mathbf{Q}$ and the action that maximizes the weighted sum is chosen.

Fig. \ref{CAN} shows the architecture of CAN and the action selection of an agent. In brief, the CAN is the decentralized execution part of the CTDE framework. The agents interact with the environment by selecting respective optimal action through the $\mathbf{Q}\left(\tau,\cdot,\boldsymbol{\omega}\right)$ generated by partial observations:
\begin{equation}
  \label{argmaxq}
  a=\underset{a}{\operatorname{argmax}} \; \boldsymbol{\omega}^{\mathsf{T}} \mathbf{Q}\left(\tau,a,\boldsymbol{\omega}\right).
\end{equation}
A mixing network is then used to generate the estimate of the joint action value function, which is for computing the TD-error to train the CAN. This will be elaborated in following sections.

\subsection{Multi-objective Mixing Network}

To handle the non-stationarity of environment issue of multi-agent systems, the centralized training is required, meaning that the behavior of agents is evaluated using joint action-value functions based on global observations.

We propose Multi-objective Mixing Network (MOMN), which is an improvement of the current MARL method, inspired by the works of VDN and QMIX.
The MOMN is a mixing network with parallel architecture for producing $Q^{tot}$ values of multiple objectives. Specifically, the input to the parallel network is the Q-vectors for n agents, i.e., $ \left\{ \mathbf{Q}^{1},\mathbf{Q}^{2},\ldots,\mathbf{Q}^{n} \right\} $, each of which consists of $m$ Q-values corresponding to $m$ objectives.
To address the multi-objective decision-making problem, MOMN is divided internally into multiple independent parallel tracks, each of which contains two neural network layers.
The Q-vectors of $n$ agents are reorganized based on objectives, where the $Q$ values corresponding to a certain objective are combined and fed into a certain MOMN track.
Finally, the outputs of each MOMN track are concatenated together to produce the $\mathbf{Q}^{tot}$, which is a vector consisting of the $Q^{tot}$ of $m$ objectives, i.e.,
\begin{equation}
  \mathbf{Q}^{tot}=\left[Q_{obj_1}^{tot},Q_{obj_2}^{tot},\ldots, Q_{obj_m}^{tot}\right],
\end{equation}
representing the joint action-values of all objectives.

MOMN must mixes all the agent action-value vectors $ \left[ \mathbf{Q}^{1},\mathbf{Q}^{2},\ldots,\mathbf{Q}^{n} \right] $ into a $\mathbf{Q}^{tot}$ and satisfy the monotonicity constraint \eqref{qmix}. To do so, several hypernetworks are used to produce weights and biases for  MLP layers of the mixing network, which is the same as the QMIX method.
For each neural network layer, two hypernetworks are used to generate its parameters. One is used for the weights and the other for the biases. Each MOMN track has two MLP layers, so there are four hypernetworks for one track.
Each hypernetwork that produces weights consists of a single linear layer and uses an absolute value activation function to ensure that the outputs are non-negative. This guarantees that the output of the mixing network satisfies the monotonicity constraint. The hypernetwork that produces the biases does not need an absolute value activation function since the biases has no non-negativity constraint. For the biases of the final layer of each track, a two-layer hypernetwork is used along with a ReLU activation function.
All hypernetworks take the global state $s$ as input, which ensures that the mixing network can utilize the global observation information.

\begin{figure}[t]
  \centering
  \includegraphics[width=0.48\textwidth]{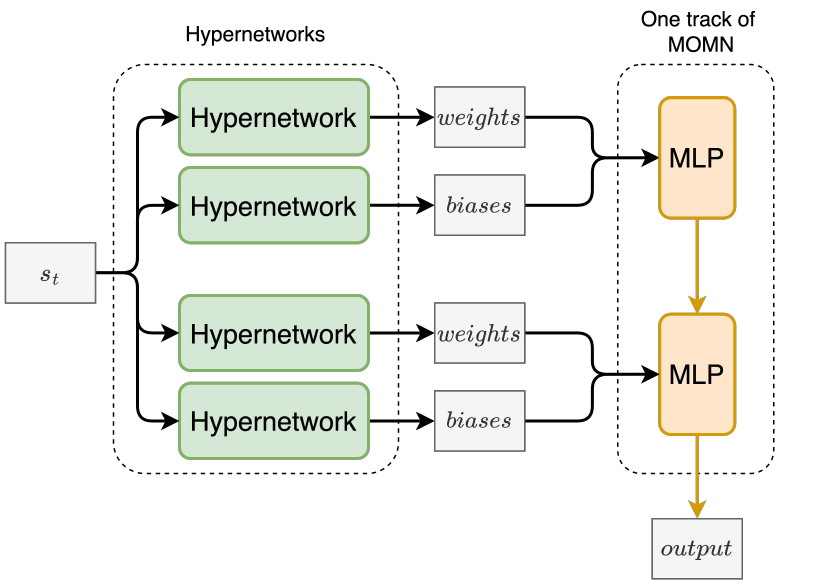}
  \caption{The connection between hypernetworks and one MOMN track. For each neural network layer of each MOMN track, two hypernetworks are used to generate its parameters. One is used for the weights and another for the biases.}  
  \label{hypernetworks}
\end{figure}

In Fig. \ref{MOMN}, it demonstrates the architecture of MOMN and the connection between CANs and MOMN. 
Fig. \ref{hypernetworks} shows the connection between hypernetworks and one MOMN track.  
In brief, MOMN takes the outputs of CANs as input and outputs the joint action-value vector $\mathbf{Q}^{tot}$. Also, MOMN satisfies the monotonicity constraint such that:
\begin{flalign}
  \underset{\boldsymbol{a}}{\operatorname{argmax}} \; \boldsymbol{\omega}^{\mathsf{T}} \mathbf{Q}^{tot}
  \left( \boldsymbol{\tau}, \boldsymbol{a}, \boldsymbol{\omega} \right)  \nonumber 
\end{flalign}
\begin{flalign}
  \label{argmaxq_mon}
  = \left(\begin{array}{c}
  \operatorname{argmax}_{a^{1}} \boldsymbol{\omega}^{\mathsf{T}} \mathbf{Q}^{1}\left( \tau^{1}, a^{1}, \boldsymbol{\omega} \right) \\
  \vdots \\
  \operatorname{argmax}_{a^{n}} \boldsymbol{\omega}^{\mathsf{T}} \mathbf{Q}^{n}\left( \tau^{n}, a^{n}, \boldsymbol{\omega} \right)
  \end{array}\right).
\end{flalign}
Therefore, the agents' actions can be selected separately by \eqref{argmaxq}, while the entire network can be trained in a centralized way.

\begin{algorithm*}[ht]
  \caption{Training procedure for MO-MIX}
  \label{moalg}
  \begin{algorithmic}

    \STATE Initialize the evaluation network parameters $\theta$, the target network parameters $\theta^{-}$, the replay buffer $\mathcal{D}$, the training batch size $N_{b}$, the preference sampling space $\Omega$.
  \FOR{each training episode}
    \STATE \textbf{Interaction phase:} 
    \STATE Sample a preference $ \boldsymbol{\omega} \sim \Omega $.
    \FOR{each interaction episode}

      \FOR{$t=1$ to $T$}
        \STATE Obtain the global state $s_{t}$ and partial observations for all agents $\mathbf{o}_{t}=\left\{ o_{t}^{1}, \ldots, o_{t}^{n} \right\}$.
        \STATE Each agent $i$ selects an action based on the $\epsilon$-greedy policy:\\
        $a^{i}_{t} =
        \begin{cases}
          \underset{a}{\operatorname{argmax}} \; \boldsymbol{\omega}^{\mathsf{T}} \mathbf{Q}^{i}
          \left( \boldsymbol{\tau}^{i}_{t}, a, \boldsymbol{\omega} ; \theta \right), & \text { with probability } 1-\epsilon  \\
          \text {random action}, & \text { with probability } \epsilon
        \end{cases}$
        \STATE Execute the joint action $ \boldsymbol{a}=\left\{ a_{t}^{1}, \ldots, a_{t}^{n} \right\} $ in the environment.
        \STATE Get the reward vector $\mathbf{r}$, the next global state $s_{t+1}$ and the next partial observation $\mathbf{o}_{t+1}$.

      \ENDFOR
      \STATE Store all transitions $\left( s_{t}, \mathbf{o}_{t}, \mathbf{a}_{t}, \mathbf{r}_{t}, s_{t+1}, \mathbf{o}_{t+1}, \boldsymbol{\omega} \right)$ of one episode in the experience replay $\mathcal{D}$.
    \ENDFOR
    \STATE \textbf{Update phase:} 
    \STATE Sample a batch from $\mathcal{D}$, consisting of $N_{b}$ transitions.
    \STATE Sample $N_{\omega}$ additional preferences $\boldsymbol{\omega}^{\prime}$.
    \STATE Calculate the TD-target according to \eqref{y}.
    \STATE Calculate the loss according to the loss function \eqref{loss} and update the evaluation network $\theta$.
    \STATE Periodically update the target network: $ \theta^{-} \leftarrow \theta $.

  \ENDFOR

  \end{algorithmic}
\end{algorithm*}

\subsection{The Exploration Guide Approach}
\label{The Exploration Guide Approach}

In the interaction phase, the optimal action of an agent is defined by \eqref{argmaxq}. That is, the vector $\mathbf{Q}$ is scalarized using the current weight $\boldsymbol{\omega}$ of preference, and the action is selected based on the scalarized $Q$ value. The weight $\boldsymbol{\omega}$ indicates the priority given to different objectives, which not only determines the evaluation criteria of the agent's behavior, but also the exploration direction of the algorithm. However, weighting the objectives does not guarantee a final policy that matches the preference. Even if a constant $\boldsymbol{\omega}$ is used throughout the training process, which effectively converts the problem into a single-objective one, there is no guarantee that the trained policy will be optimal at that weight. In the single-policy methods, the weights of each objective may require careful adjustment to obtain a satisfied policy. This is because some parts of the objective space are more difficult to achieve than others. For example, consider a task of training a walking robot with two objectives: speed and energy efficiency. If energy saving is valued, then the robot can simply stay put, or maintain a very slow speed. However, if both high speed and low energy consumption are required, the robot must carefully trade-off between the two objectives, which takes time to train and can easily converge to a more attainable suboptimal policy.

This issue also exists in MORL and is reflected in the uniformity of the final non-dominated set. We propose an exploration guide approach to alleviate this problem and improve the uniformity of final solutions. Specifically, a non-dominated set is maintained, which contains all the non-dominated solutions found so far. During the training phase, each episode samples a $\boldsymbol{\omega}$ which is used as input to the network. We divide the entire preference space into multiple subspaces and adjust the sampling probabilities of different subspaces according to the current distribution of non-dominated solutions in the objective space.
In practice, a preference vector is an angular vector with an angle range of 0-90 degrees, so we divide the space into four parts uniformly according to the angles. 
If the solutions in a certain subspace are more sparse, the sampling probability of the preferences in it will be increased. This allows weights in subspaces with poorer performance to be sampled and trained more times. The non-dominated set is reset periodically to ensure that it reflects the current quality of the policies.

\subsection{MO-MIX Algorithm}

MO-MIX belongs to the CTDE framework and is trained end-to-end using joint action values. It is also a temporal difference algorithm \cite{sutton2018reinforcement}.  The procedure of the algorithm is shown in Algorithm \ref{moalg}.

Specifically, the training process is divided into an interaction phase and an update phase. In the interaction phase, each agent selects actions based on their respective partial observations, which follows the $\epsilon$-greedy policy. 
At the beginning of each episode, a preference vector $\boldsymbol{\omega}$ is randomly sampled and used as part of the input to the MO-MIX, as described in detail in Section \ref{The Exploration Guide Approach}.  
The preference vector $\omega$ is an m-dimensional vector. Each dimension is a weight between 0 and 1, indicating the importance of one objective. In Algorithm \ref{moalg}, we use a discretized preference sampling space $\Omega$ with a sampling interval of 0.0125, which is $ \left\{ \left[ 0, 1 \right], \left[ 0.0125, 0.9875 \right] , \left[ 0.0250, 0.9750 \right] , \dots , \left[ 1, 0 \right]  \right\}$.
Then, the joint action $\boldsymbol{a}$ is executed and the environment returns a reward vector and information about the next state. An experience replay is used to store the off-policy data. At each time step, the global state, partial observations, joint action, reward vector, next global state and partial observations, and preference are stored.

In the update phase, 
we sample a batch of transitions from the replay buffer, and the preference vector is also sampled at this time. 
An evaluation network $\theta$ and a target network $\theta^{-}$ are used to increase the stability of the algorithm. The target network is a slow updating network, which is periodically synchronized with the evaluation network. Training data are sampled from the experience replay and the TD-target and loss are calculated using the target network. We also refer to the envelope value update method proposed by Yang et al. \cite{yang2019generalized}. 
Their experiments have shown that the method has good performance in the single-agent setting.
Some additional preferences $ W=\left\{ \boldsymbol{\omega}^{\prime}_{1}, \ldots, \boldsymbol{\omega}^{\prime}_{N_{\omega}} \right\} $ are sampled when computing the TD-target, and they are used to discover the potential best update target, thus improving the efficiency of learning.
The TD-target $\mathbf{y}$ can be calculated as follows:
\begin{equation}
  \label{y}
  \mathbf{y}=
  \mathbf{r} + \gamma \arg _{\mathbf{Q}} \max _{\substack{\boldsymbol{a} \in \mathcal{A}^{N} \\ \boldsymbol{\omega}^{\prime} \in W}}
  \boldsymbol{\omega}^{\mathsf{T}} \mathbf{Q}^{tot}\left( \boldsymbol{\tau}_{next}, \boldsymbol{a}, \boldsymbol{\omega}^{\prime} ; \theta^{-} \right).
\end{equation}
Then the loss function of the evaluation network $\theta$ is defined as follows:
\begin{equation}
  \label{loss}
  L(\theta)=\mathbb{E}_{\boldsymbol{\tau}, \boldsymbol{a}, \boldsymbol{\omega}}
  \left[ \left|
    \boldsymbol{\omega}^{\mathsf{T}} \mathbf{y}-\boldsymbol{\omega}^{\mathsf{T}} \mathbf{Q}^{tot}(\boldsymbol{\tau}, \boldsymbol{a}, \boldsymbol{\omega} ; \theta)
  \right| \right].
\end{equation}
The evaluation network $\theta$ is updated at every training step using \eqref{loss} and its parameters are copied to the target network $\theta^{-}$ periodically.

It is necessary to give some discussion about the link between Envelope MOQ-Learning and our method. Envelope MOQ-Learning is an advanced MORL algorithm proposed by Yang et al. \cite{yang2019generalized}. They defined the multi-objective optimization operator $\mathcal{T}$ as follows:  
\begin{multline}
  \label{envelope}
  (\mathcal{T} \mathbf{Q})(s, a, \boldsymbol{\omega}):=\mathbf{r}(s, a)+ \\
  \gamma \mathbb{E}_{s^{\prime} \sim \mathcal{P}(\cdot \mid s, a)}\arg _Q \sup _{a \in \mathcal{A}, \boldsymbol{\omega}^{\prime} \in \Omega} \boldsymbol{\omega}^{\top} \mathbf{Q}\left(s, a, \boldsymbol{\omega}^{\prime}\right) .
\end{multline}
It is proved that iteratively applying the optimization operator $\mathcal{T}$ on any multi-objective value function will terminate with a function $\mathbf{Q}$ which is equivalent to the optimal value function $\mathbf{Q}^{*}$.  

We extend the optimization operator \eqref{envelope} to the multi-agent setting, and design the TD-target \eqref{y} for multi-objective multi-agent RL. The theoretical analysis of Envelope MOQ-Learning still holds. As mentioned previously, MOMN outputs the joint action-value function $\mathbf{Q}^{tot}$ and then computes the TD-target and loss using \eqref{y} and \eqref{loss}, respectively. This is iteratively applying the optimization operator $\mathcal{T}$ on $\mathbf{Q}^{tot}$, which means that  it theoretically converges to the optimal joint action-value function $\mathbf{Q}^{tot*}$. Also, since the monotonicity condition \eqref{argmaxq_mon} is satisfied, the optimal actions chosen by multiple agents according to the partial value function is equivalent to the actions chosen according to the joint value function $\mathbf{Q}^{tot}$.  

The key difficulty of the multi-agent multi-objective setting is how to ensure the efficiency and stability of the algorithm. There are two key issues in the considered problem, i.e., ``multi-objective" and ``multi-agent", which greatly increase the problem complexity. In the proposed method, we use mixing network with parallel architecture where each track focuses on one objective, which helps to reduce the complexity and allows the algorithm to learn efficiently. Also, we introduce an exploration guide approach based on the current non-dominated set, which helps to improve the stability and performance of the algorithm.

\section{Experiments}
\label{sec5}


For multi-objective multi-agent RL, there is no universal benchmark. To verify the effectiveness of the proposed method, modifications must be made based on existing environments. We conduct experiments on the OpenAI's multi-agent particle environments (MPE)\cite{lillicrap2016continuous, mordatch2017emergence} and the StarCraft multi-agent challenge (SMAC)\cite{samvelyan2019starcraft} environments.

\subsection{Simulation Environment}

The main experimental results are obtained in the MPE, which is a widely used RL benchmark. Further, the algorithm is also tested in SMAC, a more difficult and dynamic environment.

\subsubsection{Settings for MPE}

Our experiments use the Simple Spread environment, which is part of the MPE benchmark. The environment contains a multi-agent cooperative navigation task that requires several agents to cooperate and cover all landmarks as possible. 

We use the PettingZoo \cite{terry2020pettingzoo} version of the MPE. The action space of each agent is discrete, and the agent can choose between 4 basic directions of action or do nothing. This means that the discrete action space is:
\begin{equation}
  {\mathcal{A}} = \left\{
  \begin{aligned}
    &\text{no\_action},\\
    &\text{move\_left},\\
    &\text{move\_right},\\
    &\text{move\_down},\\
    &\text{move\_up}.
  \end{aligned} \right.
\end{equation}
The global state of the environment is $ \mathbf{s} = \left[ \mathbf{x}_{1}, \ldots, \mathbf{x}_{\left(N+M\right)} \right] $, where $\mathbf{x}$ is the physical state of all the $N$ agents and $M$ landmarks, including velocity, location coordinates and type. The observation of each agent is  $ \mathbf{o} = \left[ {}_i\mathbf{x}_{1}, \ldots, {}_i\mathbf{x}_{\left(N+M\right)} \right] $, where ${}_i\mathbf{x}$ represents the observation obtained in the reference frame of agent $i$, such as relative velocity and distance.
Specifically, the physical information of an agent is defined as $ \mathbf{x} = \left[  \mathbf{p}, \dot{\mathbf{p}}, \mathbf{d}  \right] $, where $\mathbf{p}$ is the position and $\dot{\mathbf{p}}$ is the velocity of $\mathbf{p}$. $\mathbf{d}$ is the type, such as color and shape. The transformation of the physical state follows:
\begin{equation}
  \mathbf{x}_{i}^{t}
  =\left[\begin{array}{l}
  \mathbf{p} \\
  \dot{\mathbf{p}}
  \end{array}\right]_{i}^{t}
  =\left[\begin{array}{c}
  \mathbf{p}+\dot{\mathbf{p}} \Delta t \\
  \mu \dot{\mathbf{p}}+\left(\mathbf{u}+\mathbf{f}\left(\mathbf{x}_{1}, \ldots, \mathbf{x}_{N}\right)\right) \Delta t
  \end{array}\right]_{i}^{t-1},
\end{equation}
where $\Delta t$ is the timestep interval. $\mathbf{u}$ is the velocity currently selected by the agent, including the speed and direction. $\mathbf{f}\left(\mathbf{x}_{1}, \ldots, \mathbf{x}_{N}\right)$ is the physical forces between the agents and the obstacles. $\mu$ is a damping coefficient. In the observation of agent $i$, the relative position ${}_i\mathbf{p}_j$ is calculated as follows:
\begin{equation}
  {}_i\mathbf{p}_{j}=\mathbf{R}_{i}\left(\mathbf{p}_{j}-\mathbf{p}_{i}\right),
\end{equation}
where $\mathbf{R}_{i}$ is a random rotation matrix that makes each agent have a different private reference frame.

The original environment is modified to apply to the multi-objective decision-making problem. Consider a cooperative patrol task that several agents should remain around the target while not crowding in one place. To this end, we set two conflicting objectives.
\begin{enumerate}
  \item{Objective 1: The distance of the agents from the three landmarks should be as close as possible. }
  \item{Objective 2: The distance between agents should be as far as possible. }
\end{enumerate}

We use an environment containing three agents and three landmarks. At each time step, the reward obtained by agent $i$ corresponding to the objective 1 is:
\begin{equation}
  \label{obj1}
  r_{obj_1}^{i} = - \sum_{j=1}^{3} \frac{1}{3} d^{ij},
\end{equation}
where $d^{ij}$ denotes the distance from the agent $i$ to the landmark $j$.
The reward corresponding to the objective 2 is:
\begin{equation}
  \label{obj2}
  r_{obj_2}^{i} = -\alpha + \beta l^{i}_{min},
\end{equation}
where $l^{i}_{min}$ is the minimum distance between agent $i$ and other agents. The coefficients $\alpha$ and $\beta$ are used to scale the range of $r_{obj_2}^{i}$ to a similar range as $r_{obj_1}^{i}$. In our experimental settings, $\alpha=6$ and $\beta=\frac{3}{2}$.
In conclusion, the $r_{obj_1}$ will drive all agents towards a single location, while the $r_{obj_2}$ will drive the agents away from each other.
Eventually, the average of the rewards gained by all agents is used as the team reward. We tackle the multi-agent cooperation problem, which means that several agents must cooperate to optimize team rewards. Also, we use two team rewards, corresponding to two conflicting objectives, which form a multi-objective problem.

\subsubsection{Settings for SMAC}

SMAC is a representative benchmark for collaborative MARL, based on Blizzard's StarCraft II RTS game. In the original SMAC, there is only one objective, which is to win the battle.  To test the multi-objective algorithm, we set up two conflicting objectives, i.e., attack and escape. At each time step, the reward corresponding to the objective 1 is: 
\begin{equation}
  \label{r_attack}
  r_{attack} = \sum_{i=1}^{n} \left( r_{enemy}^{i}+r_{enemy\_death}^{i} \right),
\end{equation}
where $r_{enemy}^{i}$ is the damage taken by the enemy agent $i$ and $r_{enemy\_death}^{i}$ is a one-time reward for killing an enemy.
The reward corresponding to the objective 2 is:
\begin{equation}
  \label{r_escape}
  r_{escape} = - \sum_{i=1}^{n} \left( r_{ally}^{i}+r_{ally\_death}^{i} \right),
\end{equation}
where $r_{ally}^{i}$ is the damage taken by the ally agent $i$ and $r_{ally\_death}^{i}$ is a one-time penalty for an ally being killed.
Finally, we scale the reward value so that the maximum cumulative reward for an episode is 150.

\subsection{Baseline}
Currently, there is no second multi-policy MOMARL method based on deep RL. The existing methods cannot be used to solve complex MARL decision-making problems with continuous state space. Therefore, we design an outer-loop multi-objective approach for comparison. In our tests, the single-objective QMIX algorithm can perform quite well in the original MPE tasks. Also, the work of Hu et al. \cite{hu2021policy,hu2021rethinking} shows that QMIX algorithm has outperformed many newer methods and is one of the top performing MARL algorithms. Thus we choose this algorithm to construct the comparison method. It uses an outer loop to search for preferences and uses QMIX to learn policies. This approach is used as a baseline for MO-MIX. 
To ensure a fair comparison, both the baseline method and our proposed method are tested and compared in the same modified environment with rewards of (\ref{obj1}) and (\ref{obj2}). It means that the baseline method does not use a sparse reward configuration.  
The procedure of the algorithm is shown in Algorithm \ref{outerloop}.

\begin{algorithm}[H]
  \caption{Outer-loop QMIX}
  \label{outerloop}
  \begin{algorithmic}

  \STATE Initialize the preference sampling space $\Omega^{\prime}$, Number of outer cycles $N_{out}$
  \FOR{each each outer loop}
    \STATE Sample a preference $ \boldsymbol{\omega} \sim \Omega^{\prime} $ sequentially.
    \FOR{each episode}
      \STATE Agents interact with the environment.
      \STATE Scalarize the reward vector using preference $ \boldsymbol{\omega} $.
      \STATE Learning with the single-objective QMIX algorithm.
    \ENDFOR
    \STATE Store the policy corresponding to preference $ \boldsymbol{\omega} $.
  \ENDFOR

  \end{algorithmic}
\end{algorithm}

\subsection{Implementation Details}

We use a 64-dimensional GRU layer for the conditioned agent network. Before the GRU layer is an MLP layer, and after the GRU are two MLP layers. The multi-objective mixing network is consisted of two MLP layers, while the weights and biases of the two layers are generated using four hypernetworks. 
For the MPE, we use a learning rate of $5\times10^{-6}$ and train for 75,000 episodes. The network was updated using one batch of data each time, with a batch size of 64. The basic unit of data is one episode, since the GRU takes the observation history as input. The size of the replay buffer is 12000 and the discount factor $\gamma$ is 0.99.
For the SMAC, the size of the replay buffer is 20000 and the batch size is 128.

The $\epsilon$-greedy policy was used to increase the exploration of the agents. During training, each agent randomly selects actions with probability $\epsilon$ while the value of $\epsilon$ is linearly annealed from 1.0. Considering that the network requires to generalize over the entire preference space, more exploration is required. We reset the $\epsilon$ after some training steps. One preference $\boldsymbol{\omega}$ is sampled at the beginning of each episode, which allows the trained network to adapt to different preferences. We adjust the sampling probability of the preferences using the method introduced in Section \ref{The Exploration Guide Approach}. To ensure the stability of the training, we use a long update period, i.e., the target network is updated every 3500 steps. The update cycle will gradually increase during the training process.

\subsection{Results}

Four kinds of evaluation metrics are used in our experiments: the hypervolume metric (HV) \cite{zitzler2007hypervolume}, the spacing metric, the sparsity metric, and the diversity metric.

The number of final non-dominated solutions is used as the diversity metric. A higher diversity metric means that the algorithm can find more non-dominated solutions, which is better. The diversity metric is calculated as follows:
\begin{equation}
  \operatorname{Diversity}(P)=\mid P \mid,
\end{equation}
where $\mid P \mid$ denotes the number of solutions in the non-dominated set $P$.

The hypervolume metric is the volume of the area in the objective space enclosed by the reference points and the non-dominated solutions obtained by the algorithm. The larger the HV value, the better the comprehensive performance of the algorithm. HV is a Pareto-compliant evaluation method, which means that if one solution set $P$ is better than another solution set $P^{\prime}$, then the Hypervolume metric of solution set $P$ will be greater than the Hypervolume metric of solution set $P^{\prime}$. The HV is calculated as follows:
\begin{equation}
  H V=\delta\left(\bigcup_{i=1}^{\mid P \mid} v_{i}\right),
\end{equation}
where $\delta$ denotes the Lebesgue measure, which is used to measure the volume. $\mid P \mid$ denotes the number of solutions in the non-dominated set $P$, and $v_{i}$ denotes the super volume formed by the reference point and the $i$-th solution in the solution set.

The spacing metric is the standard deviation of the Euclidean distances between consecutive points in the non-dominated set, the smaller the spacing value, the more uniform the solution set. The spacing metric is calculated as follows:
\begin{equation}
  \operatorname{Spacing}(P)=\sqrt{\frac{1}{|P|-2} \sum_{i=1}^{|P|-1}\left(\bar{d}-d_{i}\right)^{2}},
\end{equation}
where $d_{i}$ denotes the distance of the $i$-th solution to the next adjacent solution in $P$ and $\bar{d}$ denotes the mean value of all $d_{i}$.

The sparsity metric is the average square Euclidean distance between consecutive points in the non-dominated set, and the smaller the sparsity metric, the denser the solution set. The sparsity metric is calculated as follows:
\begin{equation}
  \operatorname{Sparsity}(P)=\frac{1}{|P|-1} \sum_{i=1}^{|P|-1}\left(d_{i}\right)^{2}.
\end{equation}

\begin{figure}[t]
  \centering
  \includegraphics[width=0.49\textwidth]{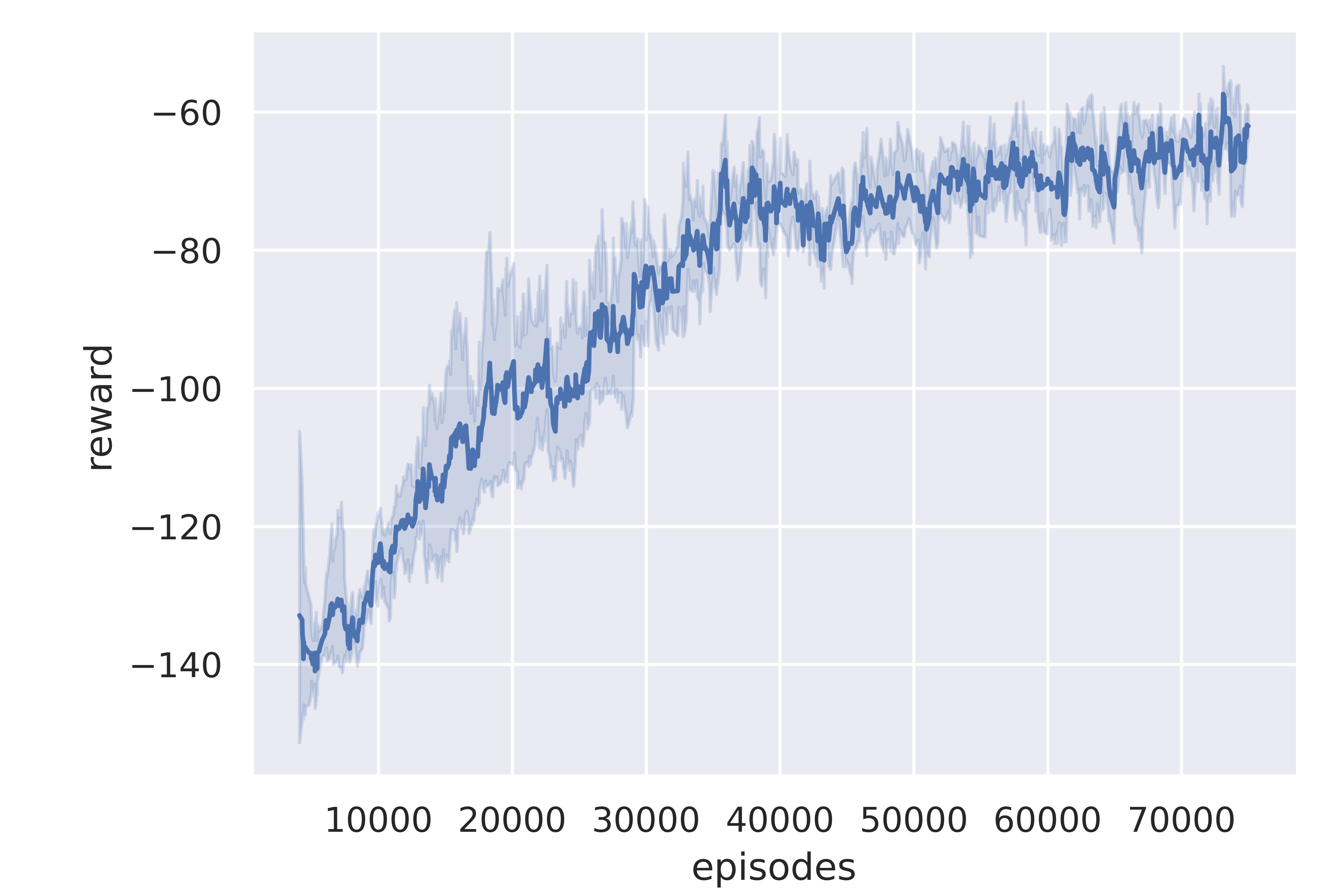}
  \caption{The average utility curve of 75000 episodes.The data are based on five independent runs of the MO-MIX algorithm and local means were estimated using a sliding average algorithm with a parameter of 0.85. The light-colored part shows the standard deviation.}
  \label{reward}
\end{figure}

\begin{figure}[t]
  \centering
  \includegraphics[width=0.49\textwidth]{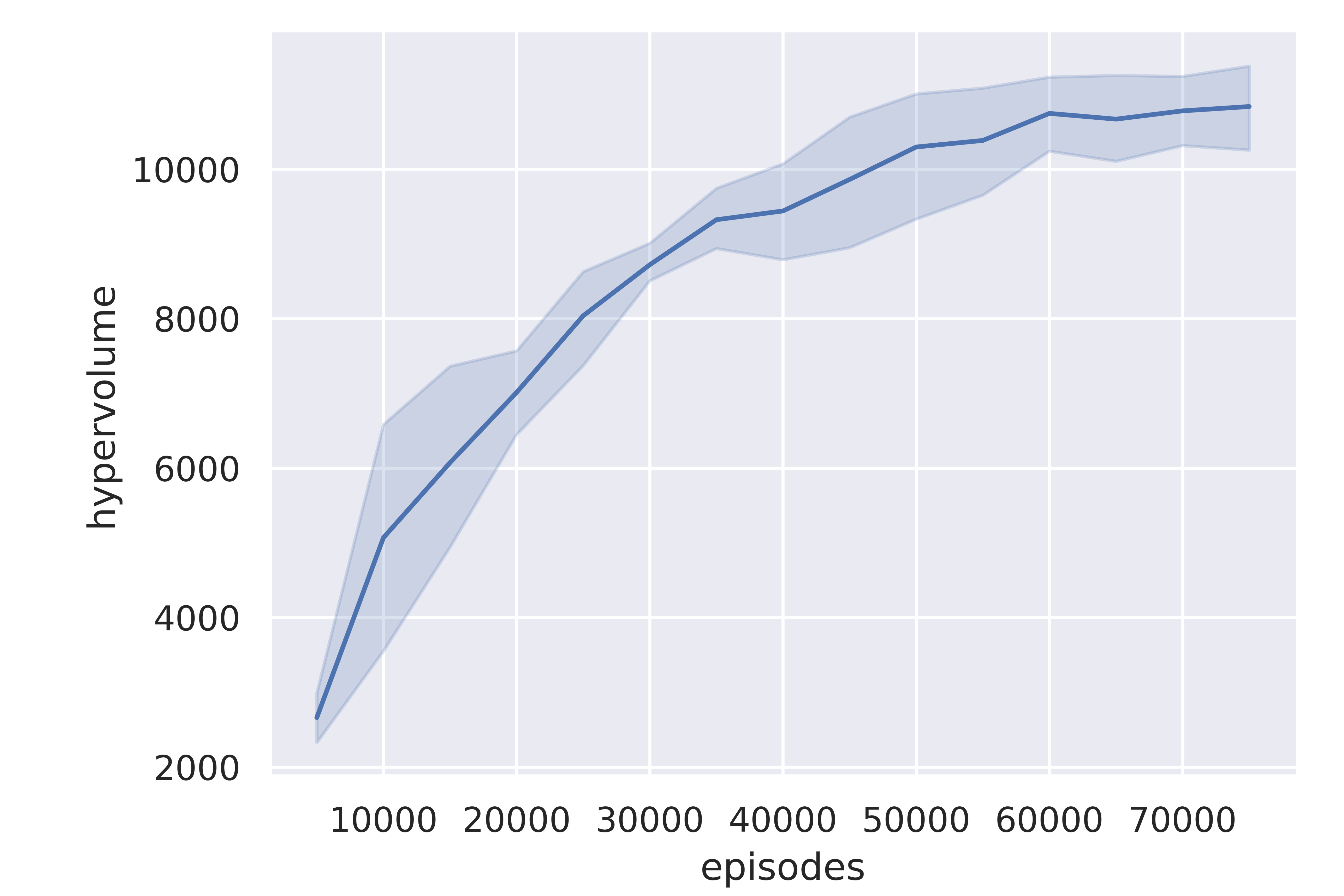}
  \caption{The hypervolume variation curve. Data are based on five independent runs of the MO-MIX algorithm, tested once every 5000 episodes. The light-colored part shows the standard deviation. }
  \label{hv}
\end{figure}

\begin{figure*}[t]
  \centering
  \includegraphics[width=0.97\textwidth]{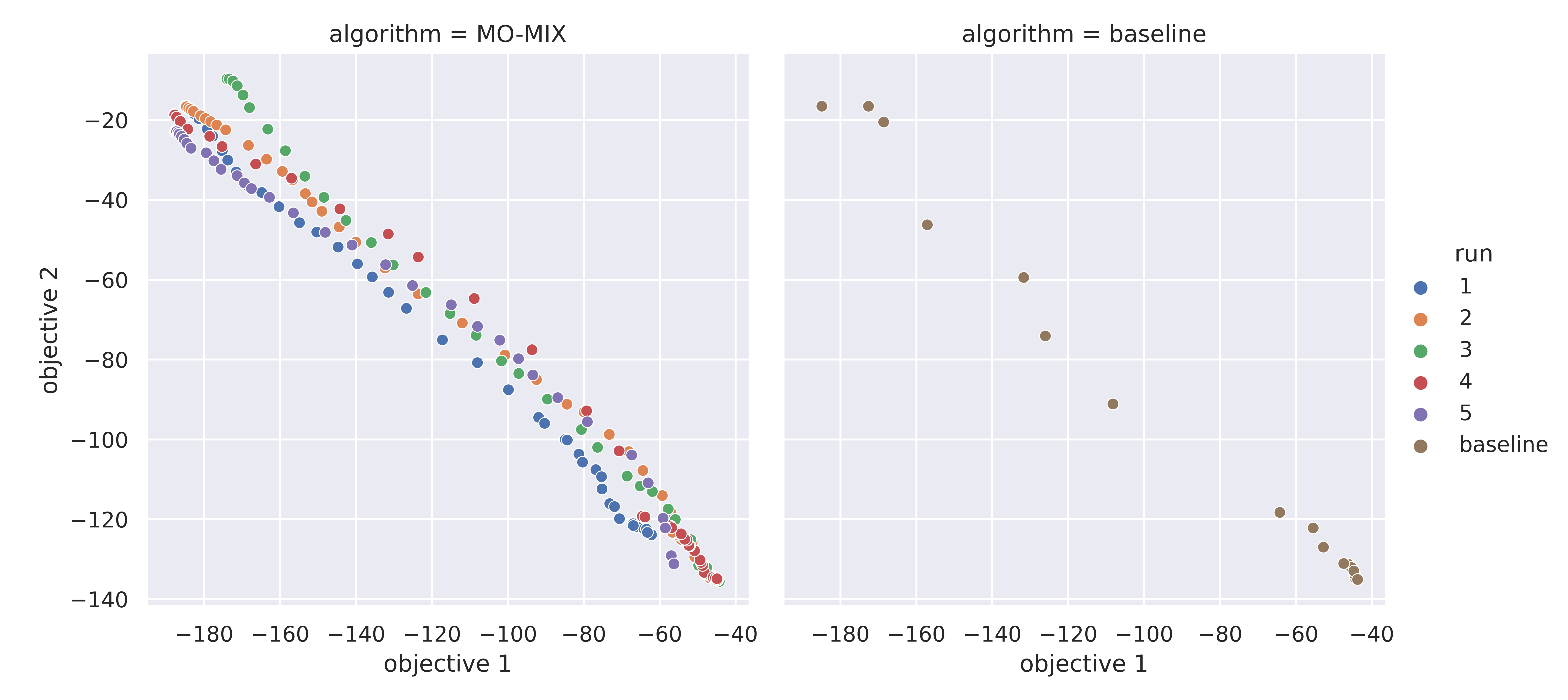}
  \caption{The final non-dominated set of the two methods. The left figure shows the results of five independent runs of the MO-MIX algorithm. Different runs are distinguished by different colors. The right figure shows the results of the outer-loop QMIX method after 41 rounds of training. All dominated solutions have been removed. }
  \label{pareto}
\end{figure*}

\begin{figure*}[t]
  \centering
  \includegraphics[width=0.97\textwidth]{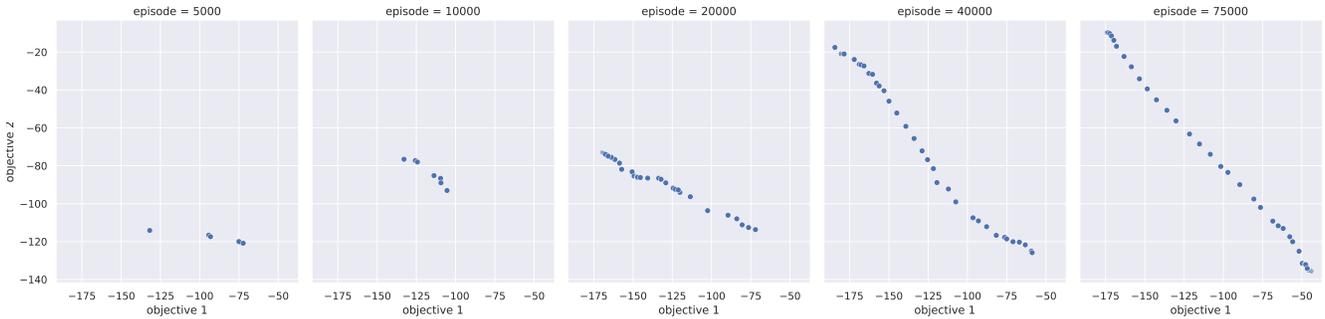}
  \caption{The movement of the non-dominated set during the entire training process. The figure shows the non-dominated set for episodes 5000, 10000, 20000, 40000, and 75000 respectively. }
  \label{pareto_history}
\end{figure*}

\subsubsection{Results on MPE}

For the MO-MIX algorithm, we show the results of five independent runs, and the sampling interval of the preferences is 0.0125. One complete training contains 75,000 episodes. For the outer-loop QMIX algorithm, we perform 41 training rounds using different preferences, each lasting 25,000 episodes. The sampling interval of the preferences is 0.025.

The average utility curve during training is shown in Fig. \ref{reward}. The utility is weighted reward that is scalarized by preferences. Rewards are evaluated every 20 episodes using a greedy policy, which means that $\epsilon = 0$. We smoothed the curve to present the average utility over a period and the variation trend. The steady growth of the utility values indicates that the learned policy is approximating the Pareto frontier.
Fig. \ref{hv} shows the variation of the hypervolume metric during the training process. We perform a complete test of the model once for each 5000 episodes. The sampling interval of preference is 0.0125. The dominated solutions are removed by performing Pareto analysis on all output policies. Finally, the HV metric is calculated for the non-dominated set. We use $\left(-200, -140\right)$ as the reference point.

\renewcommand{\arraystretch}{1.5}

\begin{table}[ht]
  \caption{Comparison of Different Algorithms on Four Evaluation Metrics\label{comparison}}
  \centering
  \setlength{\tabcolsep}{1.2mm}{
  \begin{tabular}{|c||c|c|c|}
  \hline
  Metrics & MO-MIX & {\makecell[c]{Outer-loop QMIX \\ (baseline v1)}} & {\makecell[c]{Outer-loop QMIX \\ (baseline v2)}} \\
  \hline
  Hypervolume $\uparrow$ & {\bf{11080.9224}} & 9448.3380 & 9791.7632 \\
  \hline
  Diversity $\uparrow$ & {\bf{40.40}} & 17.00 & 21.00 \\
  \hline
  Spacing $\downarrow$ & {\bf{4.2273}} & 15.8387 & 12.2401 \\
  \hline
  Sparsity $\downarrow$ & {\bf{43.6463}} & 467.3278 & 236.0842\\
  \hline
  \end{tabular}
  }
\end{table}

Fig. \ref{pareto} shows the comparison of the non-dominated set generated by the proposed MO-MIX method and the baseline method. The different independent runs are shown as points in different colors. The left figure shows the results of five independent runs of the MO-MIX algorithm. The right figure shows the results of the outer-loop QMIX method after 41 rounds of training. Dots of the same color are generated by the policies from one non-dominated set. More specifically, the x-value of a point indicates the reward obtained by a policy on objective 1. Similarly, the y-value indicates the reward obtained by the policy on objective 2. 
Multiple policies corresponding to different preferences constitute the non-dominated set. Since we tackle the multi-agent cooperation problem and agents must cooperate to gain team rewards, a policy in the non-dominated set actually contains the policies of all the agents. 
Once the non-dominated set is obtained, the user can choose an appropriate policy according to the practical needs. For example, if the user views objective 1 as much more important than objective 2, a policy should be selected in the lower right region of the diagram. A well-performing multi-objective algorithm should be able to generate high-quality non-dominated sets, which can be evaluated by the four metrics as mentioned previously. Also, Fig. \ref{pareto} provides an intuitive view showing that the MO-MIX algorithm generates better non-dominated sets than the baseline approach.  
Further, Fig. \ref{pareto_history} shows the non-dominated set at different stages of the training, which is a visual representation of the learning process.

Table \ref{comparison} shows the results of the two methods on different evaluation metrics. The hypervolume value of the non-dominated set generated by the MO-MIX method is $17.27\%$ higher than the baseline method, indicating that our method has a higher comprehensive performance. In addition, the results of MO-MIX are distinctly better than the baseline method in terms of the spacing metric, sparsity metric, and diversity metric. In terms of computational cost, the MO-MIX algorithm requires 75,000 episodes of training to reach the level shown in the figure, while the outer-loop QMIX algorithm requires 1,025,000 episodes, which exceeds the former by a factor of 13.
Further, considering that a higher sampling density of preferences may lead to better results, we retested the baseline method by changing the sampling interval of preference from 0.025 to 0.01, denoted as ``baseline v2". That is, we perform 101 training rounds using different preferences, requiring a total of 2,525,000 episodes. The experimental results are also shown in Table \ref{comparison}. The results show that baseline v2 achieves only a small performance improvement compared to baseline, but requires a much higher number of training episodes. This indicates that increasing the sampling density of preferences for the outer-loop QMIX algorithm is not an efficient approach.

\begin{figure*}[t]
  \centering
  \includegraphics[width=0.97\textwidth]{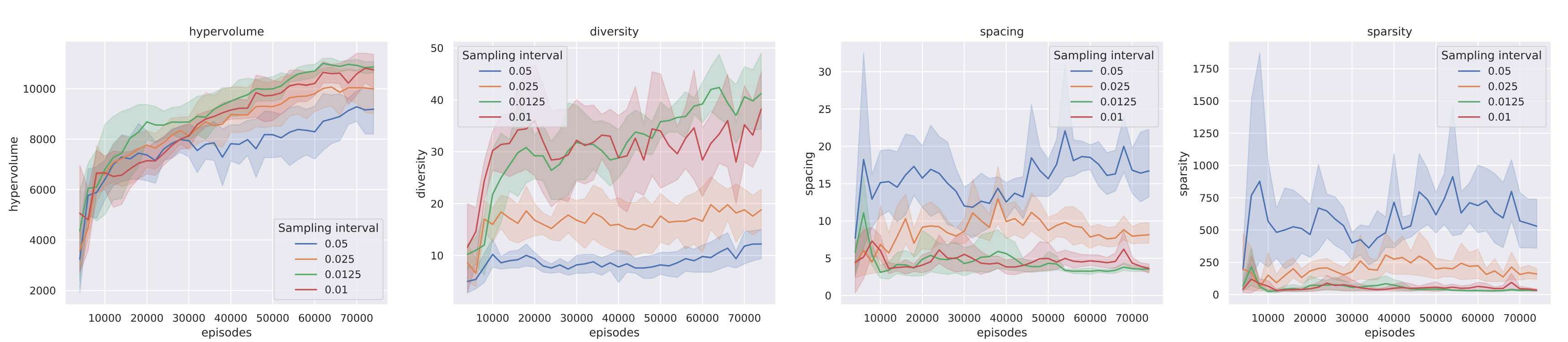}
  \caption{Results of the MO-MIX algorithm using different preference sampling intervals. The subgraphs show the variation curves of hypervolume, diversity, spacing and sparsity metrics, respectively. We plot the average performance over five experiments. The light-colored part shows the standard deviation. }  
  \label{fig_test_w}
\end{figure*}

\begin{figure*}[t]
  \centering
  \includegraphics[width=0.97\textwidth]{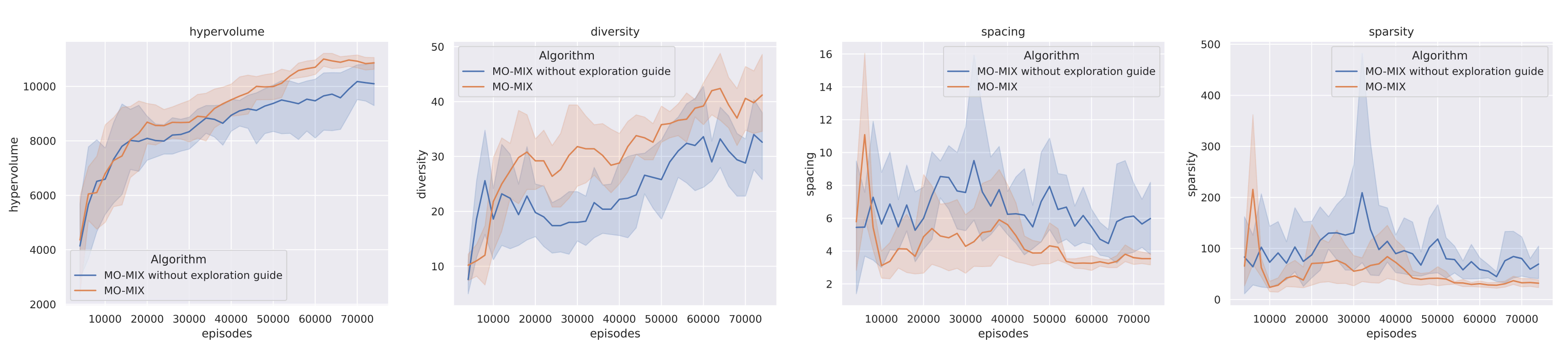}
  \caption{Results of the complete MO-MIX algorithm and an ablated version of MO-MIX in which the exploration guide part is removed. We plot the average performance over five experiments. The light-colored part shows the standard deviation.}  
  \label{fig_test_g}
\end{figure*}

Fig. \ref{fig_test_w} shows the results obtained by applying our algorithm MO-MIX with different preference sampling intervals. The subgraphs of Fig. \ref{fig_test_w} show the variation curves of hypervolume, diversity, spacing, and sparsity metrics, respectively. We can see that the algorithm performs best when the preference sampling interval is 0.0125. The evaluation metrics are clearly lower when the sampling interval is 0.05 or 0.025, which is because a large preference sampling interval leads to a sparse sample space, which also results in a low diversity of final non-dominated policies. On the other hand, changing the sampling interval from 0.0125 to 0.01 results in only a minor difference in final performance, but slows down the optimization, indicating that a sampling interval that is too small may hamper the sample efficiency. In other experiments, we choose a preference sampling interval of 0.0125 as the standard setting.

Fig. \ref{fig_test_g} shows the comparison between the complete MO-MIX algorithm and an ablated version of MO-MIX in which the exploration guide part is removed. The complete MO-MIX algorithm outperforms another on all four evaluation metrics, which shows the effectiveness of the proposed exploration guide approach.

\subsubsection{Results on SMAC}
\begin{figure*}[t]
  \centering
  \includegraphics[width=0.97\textwidth]{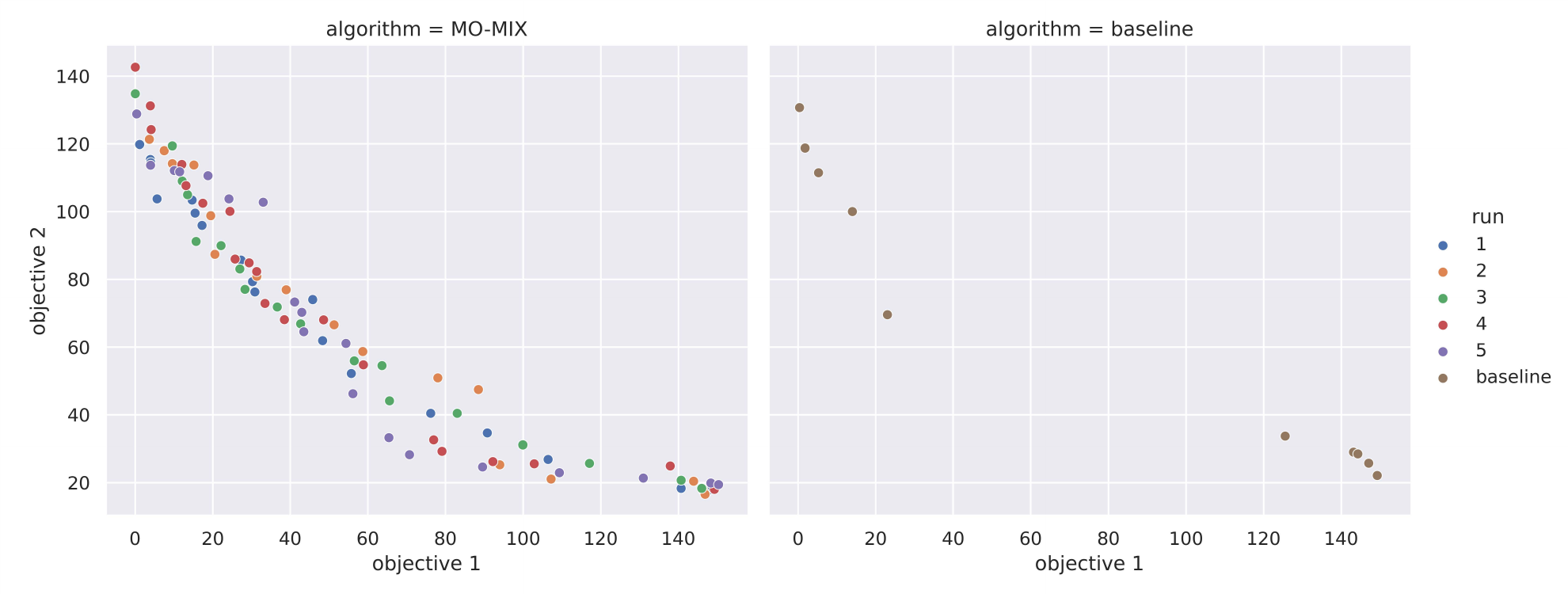}
  \caption{The final non-dominated set of the two methods, based on the SMAC-2s3z environment. The left figure shows the results of five independent runs of the MO-MIX algorithm. Different runs are distinguished by different colors. The right figure shows the results of the outer-loop QMIX method after 41 rounds of training. All dominated solutions have been removed. }\label{pareto_smac}
\end{figure*}

For the SMAC environment, the experiments are conducted in the ``2s3z" scenario. The proposed MO-MIX algorithm is trained for 5 million steps and the sampling interval of preference is 0.0125. The outer loop QMIX algorithm is trained for 41 rounds, using different preferences, each lasting 1 million steps.  
Table \ref{comparison_smac} shows the results of the two methods on different evaluation metrics. Fig. \ref{pareto_smac} shows the comparison of the non-dominated set generated by the proposed MO-MIX method and the baseline method. 

The experimental results show that the MO-MIX algorithm outperforms the baseline in the hypervolume metric and has significant advantages in the diversity, spacing and sparsity metrics. The baseline algorithm produces a non-dominated set with very low diversity and uniformity, which indicates it cannot effectively approximate the Pareto front. 
In terms of efficiency, it is note that the MO-MIX algorithm requires 5 million steps for training, while the baseline requires 41 million steps, indicating that the MO-MIX is significantly more efficient.

\begin{table}[ht]\centering
  \caption{Comparison of Different Algorithms on Four Evaluation Metrics with the SMAC environment\label{comparison_smac}}
  \centering
  \setlength{\tabcolsep}{5mm}{
  \begin{tabular}{|c||c|c|}
  \hline
  Metrics & MO-MIX & {Outer-loop QMIX} \\
  \hline
  Hypervolume $\uparrow$ & {\bf{7371.7852}} & 6226.9744 \\
  \hline
  Diversity $\uparrow$ & {\bf{18.40}} & 10.00 \\
  \hline
  Spacing $\downarrow$ & {\bf{7.8697}} & 33.5992 \\
  \hline
  Sparsity $\downarrow$ & {\bf{203.7040}} & 1509.6655 \\
  \hline
  \end{tabular}
  }
\end{table}

\subsubsection{Summary}
\label{summary}

The above experiments show the effectiveness of the proposed method. We note that the outer-loop QMIX method can reach good levels in some parts of the objective space. However, since some parts of the objective space are easier to reach than others, the outer-loop QMIX method converges to these  suboptimal solutions frequently. This leads to a final solution that may be duplicated or dominated by other similar solutions. Even though we have run the algorithm multiple times using many different preferences, we still do not end up with an ideal non-dominated set. In contrast, our MO-MIX method can explore the preference space and exploit the training trajectory more sufficiently. Specifically, some of the policies found in the random exploration phase may be very poor for the current preference, but have good performance on some other preferences. The training trajectory of other preferences can be reused when updating the policy of a particular preference. Thus, our algorithm can quickly find better policies for different preferences and approximate the Pareto frontier efficiently.

\section{Conclusion}
\label{sec6}
In this paper, we propose MO-MIX, a novel MOMARL method which aims to solve the multi-objective multi-agent cooperative decision-making problem with continuous state space. We use preferences as conditions for local action-value function estimation and generate estimates of joint action-value functions using a multi-objective mixing network with parallel architecture. Our approach can explore the preference space sufficiently and exploit training trajectories with different preferences. Thus, the algorithm can generalize over the entire preference space, and generate corresponding optimal policies based on the input preference. By feeding different preference weights to the trained model, a dense and high-quality set of policies can be obtained, which is a Pareto set approximation. In the experiments, the MO-MIX algorithm clearly outperforms the baseline in all evaluation metrics, while requiring much fewer training steps. 

Currently, our algorithm is tested in the case with two conflicting objectives. For three or more objectives, the algorithm is theoretically able to be applied. The main difference should be that the number of tracks of the mixing network needs to be set according to the number of objectives. However, further experiments are yet to be performed, which is not a trivial task. In future work, we aim to explore more complex multi-objective multi-agent problems with more difficult tasks or more diversity of objectives.


%



\ifCLASSOPTIONcompsoc
  \section*{Acknowledgments}
\else
  \section*{Acknowledgment}
\fi

This work was supported in part by the National Natural Science Foundation of China under Grants 62022094 and 61873350, the Hunan Provincial Natural Science Foundation of China under Grant 2020JJ2049, and the Innovation-Driven Project of Central South University, China under Grant 2020CX032. The authors would like to thank anonymous reviewers for their valuable comments.

\ifCLASSOPTIONcaptionsoff
  \newpage
\fi



\bibliographystyle{IEEEtran}
\bibliography{bibref}

%
\vspace{12cm}
\begin{IEEEbiography}[{\includegraphics[width=1in,height=1.25in,clip,keepaspectratio]{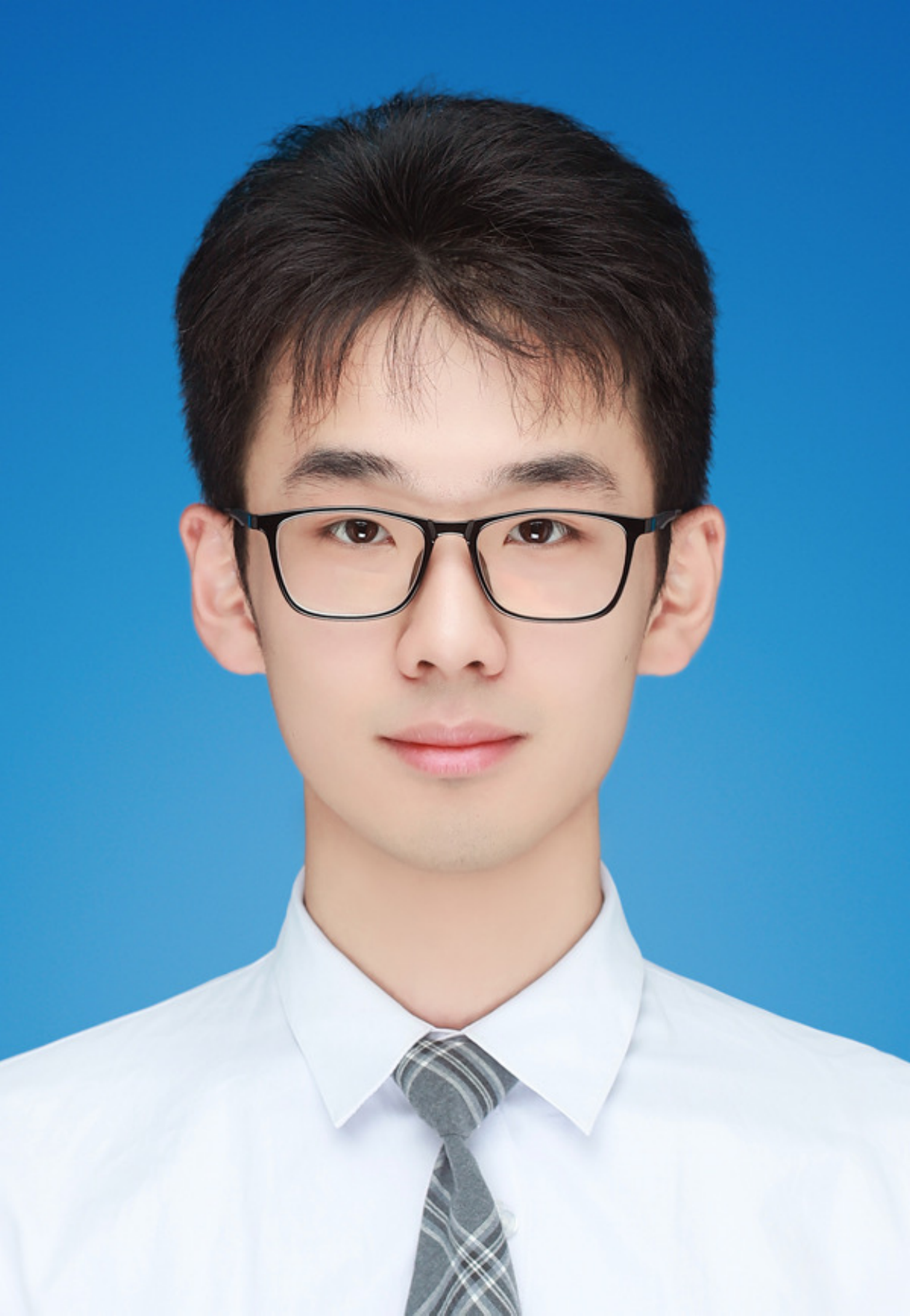}}]
  {Tianmeng Hu}
  received the B.E. degree from the School of Automation, Central South University, China, in 2021. He is currently working toward the master's degree in control science and engineering from the School of Automation, Central South University, China. His current research interests include reinforcement learning, deep learning, multi-agent systems, and multi-objective decision-making.
\end{IEEEbiography}
\vspace{-1cm}
\begin{IEEEbiography}[{\includegraphics[width=1in,height=1.25in,clip,keepaspectratio]{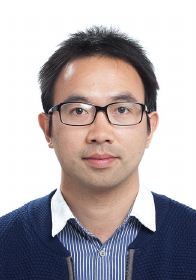}}]
  {Biao Luo}
  (Senior Member, IEEE) received the Ph.D. degree from Beihang University, Beijing, China, in 2014.
  He is currently a Professor with the School of Automation, Central South University (CSU), Changsha, China. Before joining CSU, he was an Associate Professor and Assistant Professor with the Institute of Automation, Chinese Academy of Sciences, Beijing, China, from 2014 to 2018. His current research interests include intelligent control, reinforcement learning, deep learning, decision-making.

  He serves as an Associate Editor for the \MakeUppercase{IEEE Transactions on Neural Networks and Learning Systems}, the \MakeUppercase{IEEE Transactions on Emerging Topics in Computational Intelligence}, the \textit{Artificial Intelligence Review}, the \textit{Neurocomputing}, and the \textit{Journal of Industrial and Management Optimization}. He is a Senior Member of the IEEE, and the vice chair of Adaptive Dynamic Programming and Reinforcement Learning Technical Committee, Chinese Association of Automation.
\end{IEEEbiography}
\vspace{-1cm}
\begin{IEEEbiography}[{\includegraphics[width=1in,height=1.25in,clip,keepaspectratio]{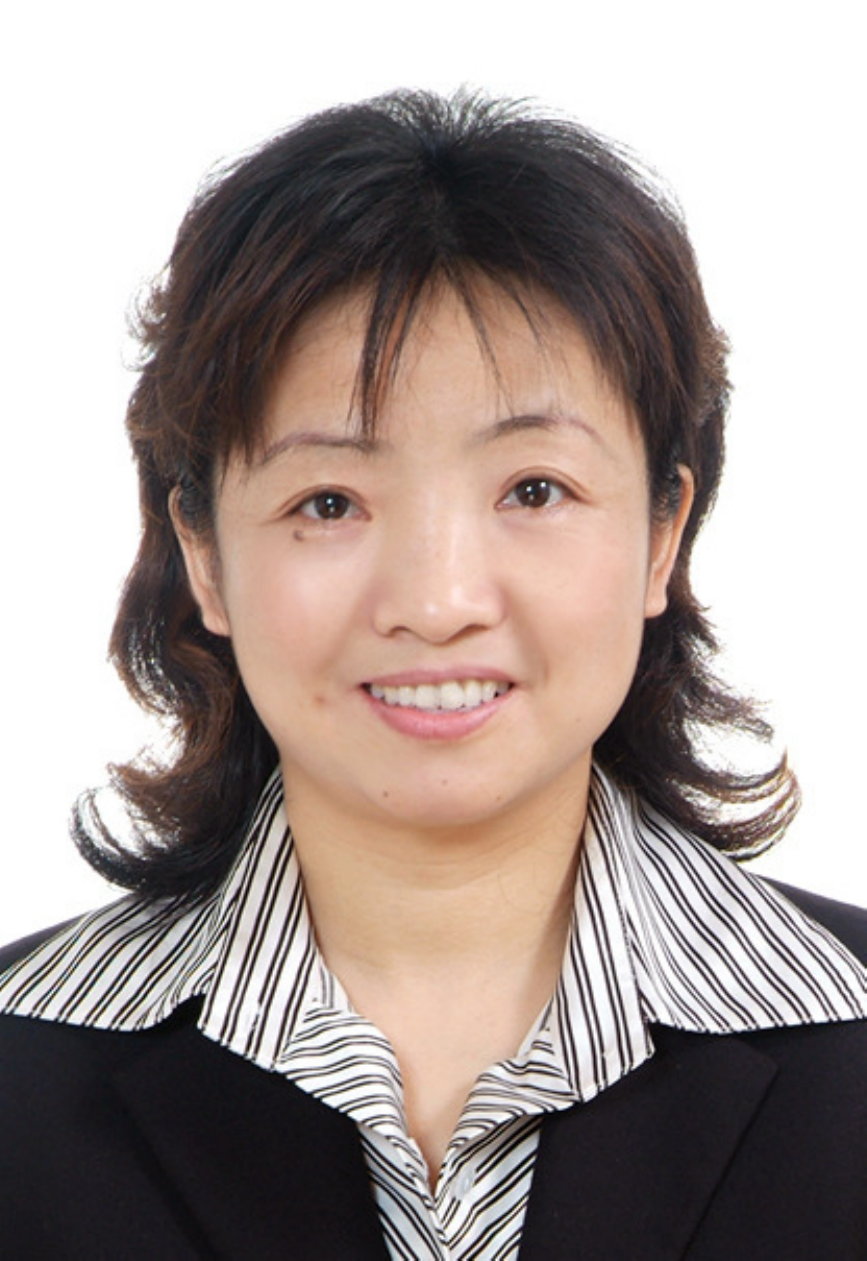}}]
  {Chunhua Yang}
  (Fellow, IEEE) received the M.S. degree in automatic control engineering and the Ph.D. degree in control science and engineering from Central South University, Changsha, China, in 1988 and 2002, respectively.

  From 1999 to 2001, she was a Visiting Professor with the University of Leuven, Leuven, Belgium. Since 1999, she has been a Full Professor with the School of Information Science and Engineering, Central South University. From 2009 to 2010, she was a Senior Visiting Scholar with the University of Western Ontario, London, ON, Canada. She is currently the Head of the School of Automation, Central South University. Her current research interests include modeling and optimal control of complex industrial processes, and intelligent control systems.
\end{IEEEbiography}
\vspace{-1cm}
\begin{IEEEbiography}[{\includegraphics[width=1in,height=1.25in,clip,keepaspectratio]{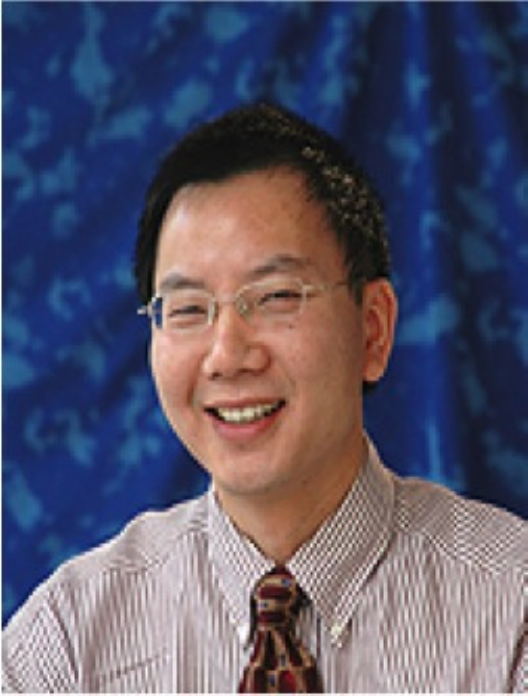}}]
  {Tingwen Huang}
  (Fellow, IEEE) is a Professor at Texas A\&M University at Qatar. He received his B.S. degree from Southwest Normal University (now Southwest University), China, 1990, his M.S. degree from Sichuan University, China, 1993, and his Ph.D. degree from Texas A\&M University, College Station, Texas, 2002. After graduated from Texas A\&M University, he worked as a Visiting Assistant Professor there. Then he joined Texas A\&M University at Qatar (TAMUQ) as an Assistant Professor in August 2003, then he was promoted to Professor in 2013. Dr. Huang's focus areas for research interests include neural networks, chaotic dynamical systems, complex networks, optimization and control.
\end{IEEEbiography}








\end{document}